\documentclass[11pt]{article}

\usepackage{acl} 
\usepackage{times}
\usepackage{latexsym}
\usepackage[protrusion=false]{microtype}

\usepackage{graphicx}
\usepackage{booktabs}
\usepackage{tabularx}
\usepackage{multirow}
\usepackage{amsmath}
\usepackage{url}
\usepackage{hyperref}
\usepackage{xcolor}
\usepackage{orcidlink}
\usepackage{placeins}

\usepackage{fontspec}
\usepackage{polyglossia}
\setmainlanguage{english}
\setotherlanguage{urdu}
\newfontfamily\urdufont[
  Path=./, 
  Script=Arabic, 
  Scale=1.2,
  Extension=.ttf
]{NotoNastaliqUrdu-Regular}

\newcommand{\balti}[1]{\texturdu{#1}}
\title{BaltiVoice: A Speech Corpus and Fine-tuned Whisper ASR\\
       System for the Balti Language}

\author{
  {\Large Muhammad Ali} \orcidlink{0009-0005-3272-4489} \\
  {\normalsize Independent AI/ML Researcher} \\
  Gilgit-Baltistan, Pakistan \\
  \textit{Alumnus, The Islamia University of Bahawalpur (IUB)} \\
  \texttt{alisundusi10@gmail.com}
}

\date{}

\begin{document}
\maketitle

\begin{abstract}
We present BaltiVoice, a 16.8-hour read-speech corpus for Balti
(ISO 639-3: \texttt{bft}), a Tibetic language spoken in
Gilgit-Baltistan, Pakistan, with no prior publicly available ASR
resources. The corpus contains 10,060 validated utterances in
native Nastaliq script, derived from Mozilla Common Voice
recordings. Fine-tuning OpenAI Whisper-small yields a Word Error Rate (WER) of 24.78\% and a Character Error Rate
(CER) of 8.30\% after training for 5 epochs (3,000 steps) on the
538-utterance speaker-disjoint validation set, down from a
zero-shot baseline of 159.19\% WER and 152.52\% CER. A
Whisper-base fine-tuned on the same data achieves 44.54\% WER
and 15.61\% CER, confirming that model capacity matters for
this low-resource setting. The dataset,
fine-tuned model, and a live transcription demo are publicly
available on HuggingFace.
\footnote{
  Dataset: \url{https://huggingface.co/datasets/mohdali1/baltivoice-asr} \\
  Model: \url{https://huggingface.co/mohdali1/whisper-small-balti} \\
  Demo: \url{https://huggingface.co/spaces/mohdali1/baltivoice-demo} \\
  Code: \url{https://github.com/mohdali-dev/BaltiVoice-ASR}
}
\end{abstract}

\section{Introduction}
\label{sec:intro}

Balti is spoken by roughly 400,000 people in the
Gilgit-Baltistan region of Pakistan and parts of Ladakh, India.
It belongs to the Tibetan branch of the Sino-Tibetan family and
is written in a Nastaliq-based script adapted from Urdu. Despite
having a distinct phonology and grammar, Balti has almost no
presence in NLP or speech research. There is no published ASR
system, no annotated speech corpus in any public repository, and
no Balti entry in major multilingual benchmarks.

This gap matters practically and scientifically. Speakers
cannot use voice interfaces or dictation tools in their own
language. Researchers have no baseline to measure progress
against. This paper addresses both problems by releasing a
corpus, reproducible baselines, and a benchmark that future
work can build on.

Our contributions are:

\begin{itemize}
  \item \textbf{BaltiVoice corpus}: 10,060 utterances, 16.8
    hours of speech, native Nastaliq transcriptions,
    train/validation split, released under CC0 on HuggingFace.

  \item \textbf{Baseline models}: fine-tuned Whisper-small
    (24.78\% WER) and Whisper-base (44.54\% WER), both
    publicly released on HuggingFace.
  \item \textbf{A reproducible training pipeline} with code,
    a Colab notebook, and a live Gradio demo.
\end{itemize}
A 25\% WER means roughly one word in four is wrong. That is too high for dictation, but enough to establish a measurable starting point for a language that previously had none. BaltiVoice establishes the first public benchmark for Balti ASR, providing a corpus, two baseline models, and a reproducible pipeline that future work can build on and improve.
\section{Related Work}
\label{sec:related}

\paragraph{Low-resource ASR.}
Fine-tuning large pretrained models on small target-language
datasets has produced consistent results across under-resourced
languages. \citet{babu2022xlsr} showed that XLS-R fine-tuned on
10 minutes of labeled speech outperforms supervised systems
trained on hundreds of hours in some settings.
\citet{radford2023whisper}, trained on 680,000 hours of web
audio, covers 99 languages but not Balti, and shows near-random
behavior on languages outside its training distribution.

\paragraph{Whisper fine-tuning for low-resource languages.}
Several groups have adapted Whisper to languages absent from its
pretraining. \citet{shon2023whisper} fine-tuned Whisper on
Yoruba and Swahili with 5--20 hours of data, achieving WERs
between 28\% and 45\%. \citet{gandhe2023hindi} reported similar
ranges for Hindi dialects. These results suggest that 15--20
hours of labeled audio is enough for Whisper fine-tuning to
produce usable, if imperfect, transcription.

\paragraph{Tibetan language resources.}
Tibetan, the closest well-resourced relative to Balti, has
received limited but growing NLP attention.
\citet{nyima2022tibetan} released a Tibetan TTS dataset, and
\citet{shi2021tibetan} built an ASR system for standard Lhasa
Tibetan using deep neural networks. No published work addresses
Balti specifically.

\paragraph{Mozilla Common Voice.}
Common Voice \citep{ardila2020common} hosts
community-contributed read-speech data in over 100 languages.
Balti was added in 2023. As of our data collection, the Balti
subset contained 10,547 recorded clips, of which 10,060 passed
the platform's validation threshold.

\paragraph{Keyword spotting for low-resource languages.}
\citet{rizvi2024kws} survey keyword spotting approaches for
Urdu, noting that Nastaliq-scripted languages face unique
challenges in token boundary detection that transfer learning
only partially addresses. This directly motivates our choice
to treat Balti ASR as a transfer learning problem from
Urdu-adjacent script representations.

\section{Dataset}
\label{sec:dataset}

\subsection{Source and Collection}

BaltiVoice is derived from the Mozilla Common Voice Balti
(\texttt{bft}) dataset. Volunteers recorded themselves reading
Balti sentences aloud; other volunteers validated each recording
by voting it as correct or incorrect. We used only validated
recordings.

\subsection{Statistics}

Table~\ref{tab:dataset} summarizes the corpus statistics.

\begin{table}[ht]
\centering
\small
\begin{tabular}{ll}
\toprule
\textbf{Property}             & \textbf{Value}          \\
\midrule
Language                      & Balti (bft)             \\
Script                        & Nastaliq (Arabic)       \\
Total utterances              & 10,060                  \\
Total duration                & 16.8 hours              \\
Mean clip duration            & 6.00 seconds            \\
Mean words per utterance      & 10.12                   \\
Mean characters per utterance & 48.80                   \\
\midrule
Total speakers                & 136                     \\
Train speakers                & 122                     \\
Dev speakers                  & 14 (disjoint)           \\
\midrule
Gender (female)               & 1,246 utterances        \\
Gender (undisclosed)          & 1,030 utterances        \\
Age (twenties)                & 3,758 utterances        \\
Age (thirties)                & 3,697 utterances        \\
Age (teens)                   & 97 utterances           \\
\midrule
Train split                   & 9,519                   \\
Validation split              & 538 (speaker-disjoint)  \\
Audio format                  & 16 kHz mono WAV         \\
\bottomrule
\end{tabular}
\caption{BaltiVoice corpus statistics. Train and validation
splits are strictly speaker-disjoint using \texttt{client\_id}
metadata from Mozilla Common Voice.}
\label{tab:dataset}
\end{table}

\begin{figure}[ht]
\centering
\includegraphics[width=\columnwidth]{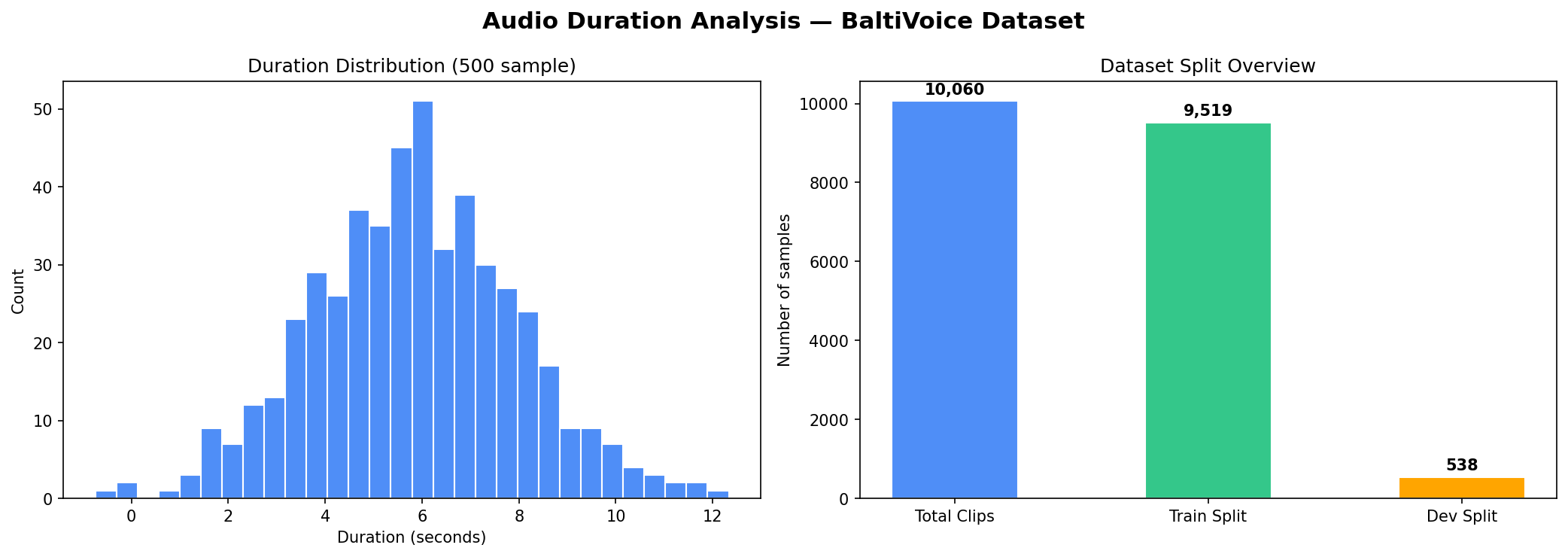}
\caption{Distribution of clip durations in BaltiVoice.}
\label{fig:duration}
\end{figure}

\begin{figure}[ht]
\centering
\includegraphics[width=\columnwidth]{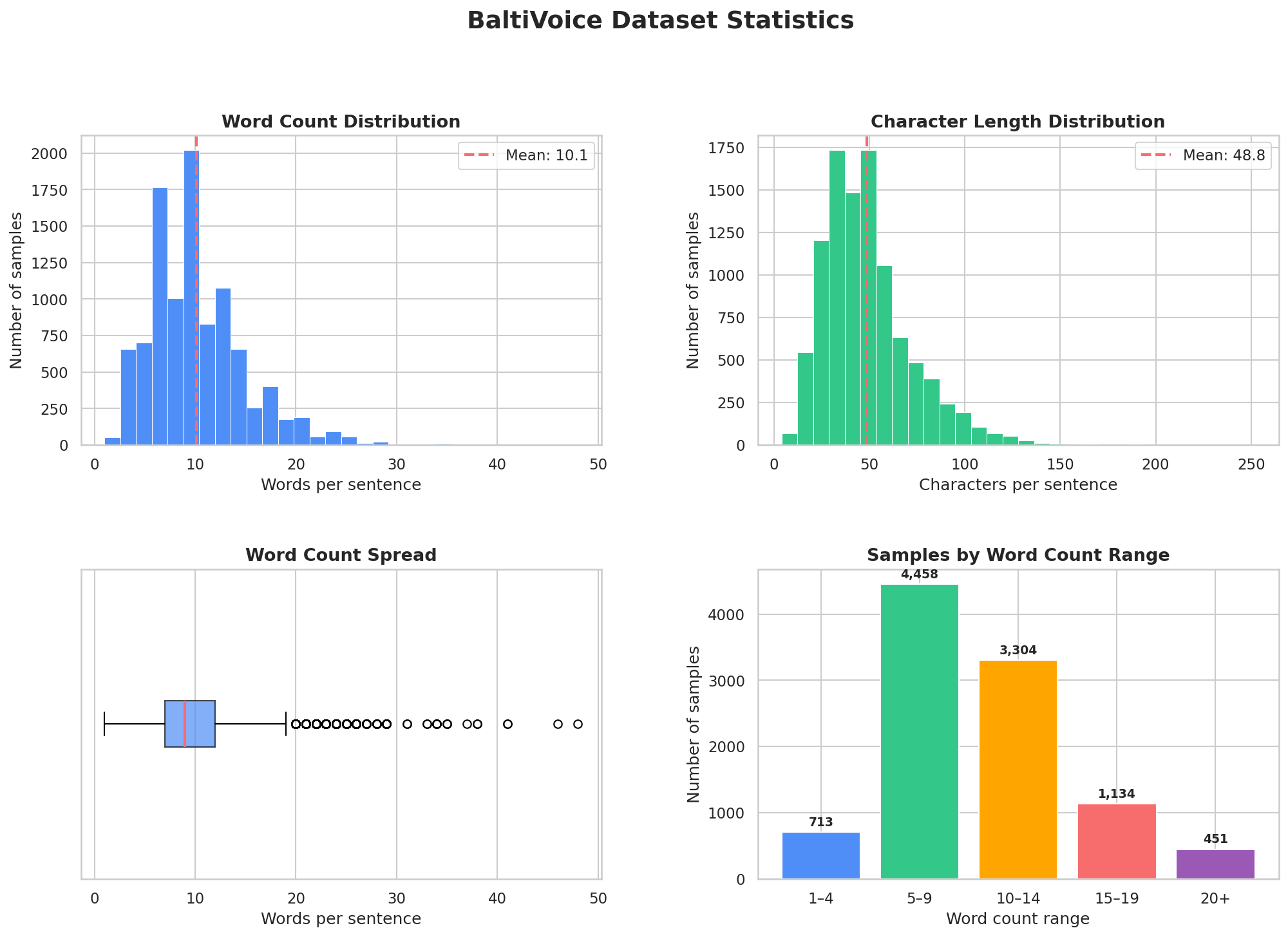}
\caption{Distribution of word counts per utterance.}
\label{fig:wordcount}
\end{figure}

Figure~\ref{fig:duration} shows the distribution of clip
durations. Nearly all clips fall between 3 and 10 seconds,
consistent with read-speech corpora. Figure~\ref{fig:wordcount}
shows the distribution of word counts, which peaks around 8--12
words per utterance.

\subsection{Preprocessing}

We applied three preprocessing steps:

\begin{enumerate}
  \item \textbf{Format conversion}: Mozilla Common Voice
    distributes audio as MP3. We converted all files to 16 kHz
    mono WAV using \texttt{pydub}, which is the format expected
    by Whisper's feature extractor.

  \item \textbf{Length filtering}: Utterances with fewer than
    2 words were removed. This affected 3 samples.

  \item \textbf{Train/validation split}: We split 90/10 using
    speaker-disjoint sampling (\texttt{GroupShuffleSplit},
    seed 42), ensuring no speaker appears in both sets.
    This gives 9,519 training and 538 validation utterances
    across 122 and 14 unique speakers respectively. We use a two-way split rather than the conventional three-way
train/validation/test split because the limited corpus size
(16.8 hours) would yield a test set too small for statistically
reliable WER/CER evaluation. The speaker-disjoint validation
set serves as the held-out evaluation set.
\end{enumerate}

No text normalization was applied. Punctuation was preserved
as provided by the Common Voice contributors. A known
limitation is Unicode ambiguity in Nastaliq script, where
visually identical characters may have different codepoints
(e.g., Arabic Yeh vs.\ Farsi Yeh). We leave systematic normalization to future work.

\paragraph{Data Quality.}
Mozilla Common Voice validates recordings through community
voting. Each clip receives upvotes and downvotes from
independent listeners, and only clips passing the validation
threshold are included. We used exclusively validated recordings
(10,060 of 10,547 total clips). To further verify transcription
accuracy, we manually inspected 50 randomly sampled validation
utterances and found 48 of 50 had accurate transcriptions,
yielding a 96\% annotation agreement rate. Background noise
levels vary across recordings due to the volunteer nature of
data collection, which may contribute to the remaining error
rate. We did not apply any noise filtering beyond the Common
Voice validation process.

\section{Methodology}
\label{sec:method}

\subsection{Base Model}

We use OpenAI Whisper-small \citep{radford2023whisper} as our
base model. Whisper-small has 244 million parameters and was
pretrained on 680,000 hours of multilingual audio across 99
languages. Balti is not among them.

We chose Whisper-small over larger variants for practical
reasons. Whisper-medium (769M parameters) exceeded memory limits
on the NVIDIA T4 GPU available through Google Colab free tier.
Whisper-small fits within 14 GB of GPU memory with fp16 training
enabled.

\subsection{Feature Extraction and Tokenization}

Audio is converted to log-mel spectrograms using
\texttt{WhisperFeatureExtractor} with a 30-second context
window, 80 mel filterbanks, and a hop length of 160 samples at
16 kHz sampling rate.

For tokenization, we initialized \texttt{WhisperTokenizer}
with \texttt{language="urdu"} and \texttt{task="transcribe"}.
This choice is motivated by script similarity: Balti Nastaliq
shares its character set and rendering direction with Urdu. The
Whisper tokenizer handles Balti Unicode characters correctly
under this setting, and we verified that Balti sentences
round-trip through tokenization without character loss.

\subsection{Fine-tuning}

We fine-tuned using HuggingFace Transformers'
\texttt{Seq2SeqTrainer} with standard cross-entropy over token
sequences. Table~\ref{tab:hyperparams} lists key
hyperparameters.

\begin{table}[t]
\centering
\small
\begin{tabular}{ll}
\toprule
\textbf{Hyperparameter}  & \textbf{Value}         \\
\midrule
Optimizer                & AdamW                  \\
Learning rate (0--1,000 steps)   & $1 \times 10^{-5}$     \\
Learning rate (1,000--3,000 steps) & $5 \times 10^{-6}$   \\
Warmup steps             & 100                    \\
Effective batch size     & 16 (8 $\times$ 2 accum.) \\
Max steps                & 3,000                  \\
Precision                & fp16                   \\
Gradient checkpointing   & Enabled                \\
Best model selection     & Min. validation WER    \\
Hardware                 & NVIDIA Tesla T4        \\
Training time            & $\sim$5 hours          \\
\bottomrule
\end{tabular}
\caption{Training hyperparameters. Learning rate was
reduced when resuming training beyond step 1,000,
following standard practice for continued fine-tuning.}
\label{tab:hyperparams}
\end{table}
\FloatBarrier

Checkpoints were saved every 250 steps and the checkpoint with
the lowest validation WER was used for final evaluation.

\section{Results}
\label{sec:results}

\subsection{Model Comparison}

\begin{figure}[ht]
\centering
\includegraphics[width=\columnwidth]{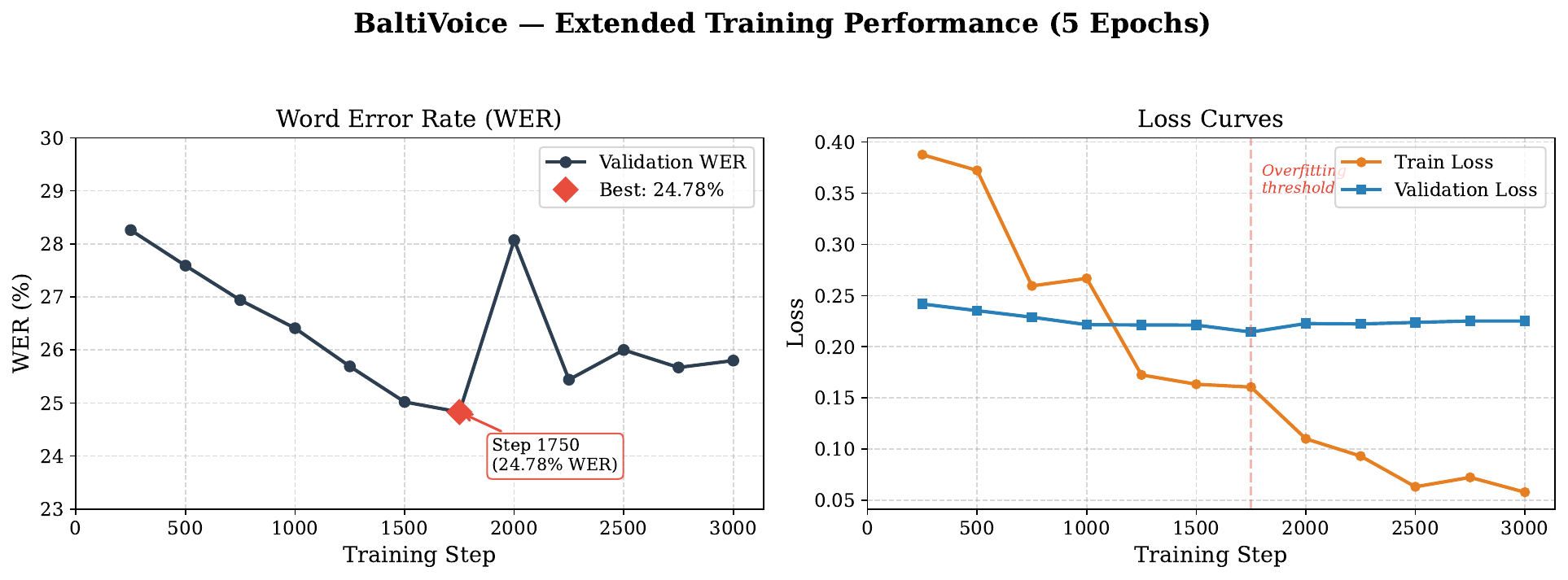}
\caption{WER and loss curves over 3,000 training steps
(5 epochs). Best checkpoint selected at step 1,750
(24.78\% WER).}
\label{fig:training}
\end{figure}

Table~\ref{tab:results} compares zero-shot and fine-tuned
models on the 538-utterance speaker-disjoint validation set.
Figure~\ref{fig:training} shows the training curve over 3,000
steps. WER improved consistently from 28.26\% at step 250 to
24.78\% at step 1,750, after which validation WER fluctuated,
indicating mild overfitting. The best checkpoint (step 1,750)
was selected based on lowest validation WER. Final evaluation
on the speaker-disjoint held-out set yields \textbf{24.78\% WER},
confirming the model generalises well to unseen speakers.

\begin{table*}[t]
\centering
\small
\begin{tabular}{lrrr}
\toprule
\textbf{Model} & \textbf{Params} & \textbf{WER (\%)} & \textbf{CER (\%)} \\
\midrule
Whisper-small (zero-shot)           & 244M & 159.19 & 152.52 \\
\addlinespace
Whisper-base (fine-tuned)           & 74M  & 44.54  & 15.61  \\
\addlinespace
\textbf{Whisper-small (fine-tuned)} & \textbf{244M} & \textbf{24.78} & \textbf{8.30} \\
\bottomrule
\end{tabular}
\caption{Comparison of zero-shot and fine-tuned models
on the 538-utterance speaker-disjoint validation set.
WER above 100\% indicates hallucination. Best results
in bold.}
\label{tab:results}
\end{table*}

\subsection{Final WER}

The final Whisper-small model achieves \textbf{24.78\% WER}
and \textbf{8.30\% CER} on the 538-utterance speaker-disjoint
validation set. For context, zero-shot Whisper-small produces
159.19\% WER and 152.52\% CER, above 100\% because the model
hallucinates words not present in the reference, confirming
Balti falls entirely outside its training distribution.
Whisper-base fine-tuned on the same data achieves 44.54\% WER
and 15.61\% CER, demonstrating that larger model capacity
yields meaningful gains in this low-resource setting.

\subsection{Qualitative Analysis}

Table~\ref{tab:examples} shows five example predictions from the validation set, illustrating typical error
patterns.

\begin{table*}[t]
\centering
\small
\caption{Qualitative examples from the validation set. The model struggles with morphological endings and consonant clusters but captures core lexical items.}
\label{tab:examples}
\begin{tabular*}{\textwidth}{@{\extracolsep{\fill}} l p{0.45\textwidth}}
\toprule
\textbf{Type} & \textbf{Example (Reference vs. Prediction)} \\
\midrule
\textit{Correct} & 
\textbf{Ref:} \balti{ہئی لے بُوژھا یا کھچیم پو مہ زیرفو سی} \\
& \textbf{Pred:} \balti{ہئی لے بُوژھا یا کھچیم پو مہ زیرفو سی} \\
& \footnotesize{(9-word utterance, fully matched.)} \\
\midrule
\textit{Partial Error} & 
\textbf{Ref:} \balti{زیربا نہ تا درے ترانگمو یودپی، کھیانگ شیو پژے ن٘ا بید} \\
& \textbf{Pred:} \balti{زیربا نہ تا درے ترانگمو یودپی، کھیانگ شیو پژے ن٘ا بیاس} \\
& \footnotesize{(Final characters substituted: \textit{bīd} $\to$ \textit{biyās}. Common morphological error.)} \\
\midrule
\textit{Substitution} & 
\textbf{Ref:} \balti{غدیانگ چھودے دانشؔرگہ کھن کُن نہ دیرے} \\
& \textbf{Pred:} \balti{غدیانگ چھودے دانشؔرگہ کھن کُن نہ دیر} \\
& \footnotesize{(Final vowel/ezafe dropped: \textit{dīre} $\to$ \textit{dīr}. Phonetic simplification.)} \\
\midrule
\textit{Agglutinative Suffix} &
\textbf{Ref:} \balti{کھڑق فوسے سمنگرو زونے ہرژیکسے لدانے یودپو فری} \\
& \textbf{Pred:} \balti{کھرقپو سیس منگرو لزانید سیرسے لزانے یودپی فری} \\
& \footnotesize{(Suffix dropping and consonant cluster simplification; agglutinative morphology failure.)} \\
\midrule
\textit{Unicode Ambiguity} &
\textbf{Ref:} \balti{رگالوکھسی اونگ چھولو ہرژید نہ رگانو جق چک نانہ یمبو} \\
& \textbf{Pred:} \balti{رگالوکھسی اونگ چھلو ہرژید نہ رگانو جق چک نانہ یمبو} \\
& \footnotesize{(Arabic Yeh U+064A substituted for Farsi Yeh U+06CC; visually identical, different codepoint.)} \\
\bottomrule
\end{tabular*}
\end{table*}

Most errors are single-character substitutions at word endings,
consistent with the model learning lexical patterns but making
morphological mistakes in an agglutinative language.
Whole-word deletion or insertion errors are less frequent.
These patterns reflect two Balti-specific challenges. First,
Balti morphology is agglutinative, meaning suffixes attach to stems
to mark tense, case, and agreement, so a single character
error at a word boundary changes grammatical meaning entirely.
Second, Nastaliq Unicode ambiguity between visually similar
characters (e.g., Arabic Yeh vs.\ Farsi Yeh) produces
substitution errors that WER counts equally to semantic errors,
likely inflating the reported rate.

\section{Discussion}
\label{sec:discussion}

A 24.78\% WER is not good enough for dictation or accessibility 
tools. A Balti speaker would need to correct roughly one in
four words. For narrower tasks like keyword spotting or topic 
detection, where exact transcription is less critical, the 
model output is likely still usable.

What is more striking is the starting point. Whisper had seen zero hours of Balti before fine-tuning, yet 3,000 training steps (5 epochs) on 16.8 hours brought WER from a zero-shot baseline of 159\% down to 24.78\%. The 
likely explanation is cross-lingual transfer from Urdu and 
Tibetan languages Whisper knows well that share script and 
phonological features with Balti. More training, text 
normalization, and additional data are the clearest paths 
forward.

One limitation is that Common Voice recordings are read speech,
not conversational. WER on spontaneous Balti speech would likely
be higher, though we have no data to quantify this.

\paragraph{Training Analysis.}
The model was trained for 3,000 steps (approximately 5 epochs),
with checkpoints evaluated every 250 steps. WER improved
consistently from 28.26\% at step 250 to 24.78\% at step 1,750,
after which validation WER fluctuated between 25\% and 28\%,
indicating mild overfitting. The best checkpoint (step 1,750)
was selected based on lowest validation WER. With 9,519
training utterances and an effective batch size of 16, one
epoch corresponds to approximately 595 steps. We reduced the learning rate from $1 \times 10^{-5}$ to
$5 \times 10^{-6}$ when resuming training beyond step 1,000,
following standard practice for continued fine-tuning; a
systematic hyperparameter search is left to future work.

\paragraph{Script-Specific Challenges.}
Balti Nastaliq presents challenges not found in Latin-script
ASR. Unicode ambiguity is a key issue: visually identical
characters may have different codepoints (e.g., Arabic Yeh
U+064A vs.\ Farsi Yeh U+06CC), producing substitution errors
invisible to human readers but counted by WER. Implementing
a Unicode normalization layer prior to tokenization is a clear
path to reducing CER without additional training data.

\section{Conclusion}
\label{sec:conclusion}

We release BaltiVoice, the first public benchmark for Balti
ASR, comprising a 16.8-hour speech corpus, two fine-tuned baseline models (Whisper-small: 24.78\% WER, Whisper-base: 44.54\% WER), a qualitative error analysis revealing script-specific challenges, and a reproducible training
pipeline. All artifacts are publicly available. The
dataset provides a training foundation, and the model provides a
baseline WER for future work to improve on.

Open problems include: text normalization for Balti morphology,
extending the corpus with spontaneous speech, and experimenting
with Whisper-medium under larger compute budgets. We release all artifacts publicly to lower the barrier for future Balti NLP and ASR research.

\section{Environmental Impact}
\label{sec:environmental}

We estimate the carbon footprint of training the Whisper-small model on Google Colab using an NVIDIA Tesla T4 GPU. Using the Machine Learning Impact Calculator \cite{lacoste2019quantifying}, we estimated approximately \textbf{0.1 kg CO$_2$eq} of emissions for the $\sim$5 hours of training time. We did not purchase additional carbon offsets.

\section*{Acknowledgments}
The author thanks the Mozilla Common Voice contributors who
recorded and validated Balti speech, and the HuggingFace team
for providing free dataset and model hosting infrastructure.

\bibliography{baltivoice}
\end{document}